\documentclass{article}       
\usepackage{graphicx}
\usepackage{amsfonts}
\usepackage{amsmath}
\usepackage{amssymb}
\usepackage{tikz}
\usetikzlibrary{arrows}
\usetikzlibrary{calc,matrix,fit}
\usepackage[bookmarks = true, colorlinks=true, linkcolor = black, citecolor = black, menucolor = black, urlcolor = black]{hyperref}

\newtheorem{theorem}{Theorem}
\newtheorem{corollary}{Corollary}
\newtheorem{definition}{Definition}
\newtheorem{remark}{Remark}

\newcommand{\Zset}{\mathbb{Z}}
\newcommand{\Z}{\mathbb{Z}}
\newcommand{\Nset}{\mathbb{N}}         

\DeclareMathOperator{\pers}{pers}
\DeclareMathOperator{\inc}{inc}

\DeclareMathOperator{\im}{Im} 
\DeclareMathOperator{\kr}{Ker} 

\newcommand{\rrdc}{\mbox{\,\(\Rightarrow\hspace{-9pt}\Rightarrow\)\,}}
\newcommand{\lrdc}{\mbox{\,\(\Leftarrow\hspace{-9pt}\Leftarrow\)\,}}
\newcommand{\lrrdc}{\mbox{\,\(\Leftarrow\hspace{-9pt}\Leftarrow\hspace{-5pt}\Rightarrow\hspace{-9pt}\Rightarrow\)\,}} 

\begin{document}

\title{Effective persistent homology of digital images}



\author{Ana Romero        \and
                Julio Rubio \and
        Francis Sergeraert
}



\maketitle

\begin{abstract}
In this paper, three Computational Topology methods (namely effective homology, persistent homology and discrete vector fields) are mixed together to produce algorithms for homological digital image processing. The algorithms have been implemented as extensions of the Kenzo system and have shown a good performance when applied on some actual images extracted from a public dataset.
\end{abstract}

\section{Introduction}
\label{sec:intro}

Computational Topology and, in particular, homological algorithms are now well established methods to study digital imaging (see, for instance, \cite{EH10}). Also in \emph{Fundamental} Algebraic Topology, effective methods have been used to compute homology and homotopy groups of abstract spaces, and even more complicated invariants (see, for example, \cite{CKMSVW12}). In this field, \emph{effective homology} has provided tools to prove the computability of theoretically defined objects. In parallel, effective homology ideas have been implemented in the \emph{Kenzo} system, allowing us to materialise the computability theorem into concrete results obtained by computer (see \cite{Kenzo}).

When effective homology methods are applied to the case of persistent homology, a well-known tool in Computational Topology, we obtain (see \cite{RHRS13}) a procedure that smoothly generalises the usual approach to persistence at least in three aspects:
\begin{enumerate}
\item our algorithm can be applied to infinite spaces,
\item we can work with integer coefficients (instead of working with coefficients over a field, as it is common in the literature, see \cite{EH08,ZC05}), and
    \item we don't loose the geometrical link with the initial space along the process of eliminating useless information (a process needed to compute the final groups); to be more precise, we can obtain the generators of the persistent homology groups expressed as cycles in the initial space.
\end{enumerate}

In the context of 2D-digital images, the two first items are not directly applicable, since they are finite objects with torsion-free homology (and then coefficients over a field capture all the homological information). The third item could be important to establish a \emph{qualitative} analysis (based on the intrinsic geometrical meaning encoded in cycles) instead of the usual \emph{quantitative} analysis (based on the dimension of homology groups and the associated \emph{barcodes}; see \cite{EH10}).

The fundamental notion in effective homology is that of (algebraic) \emph{reduction}. A reduction discards information which is useless from the homological point of view, but keeping a (functional) link with the original space (providing the information previously mentioned  in item 3). The technique proposed in this paper to get reductions for digital images is that of \emph{Discrete Vector Fields} (DVF in the sequel). We present a new algorithm to obtain a DVF from a digital image. Although the DVF is not always optimal (it is well-known that the computation of optimal DVFs is a \emph{hard} problem, ``hard'' in the technical sense of complexity theory), it has performed remarkably in application cases.

Then, this algorithm is adapted to the \emph{filtered} case, allowing us to compute reductions which are useful for determining persistent homology. The algorithm is not really adapted, but directly applied to some diagonal blocks of a matrix (where blocks are defined by filtration indexes). In that way, we get worst reduction levels  than in the global case (when applying the algorithm to the whole matrix), but with the benefits of reaching quite directly the persistent homology information. Other positive features of our algorithm is that the different DVF components can be gathered together simply by concatenation (without paying attention to a technical constraint, that of \emph{admissability}, that complicates the global algorithm), performing efficiently and opening the possibility of a parallel processing. Looking at our results in a global way, they generalise those of \cite{MN13}.

Our algorithms have been implemented as extensions of the Kenzo system~\cite{Kenzo}, and applied to some digital images. After some application to \emph{artificial} examples (constructed by us), we looked for public datasets of digital images. We found a repository of actual \emph{fingerprints}, and applied our programs to some of them, obtaining a remarkable performance with respect to the number of elements discarded to get its persistent homology groups. It is important to stress that we choose this fingerprint database as benchmark for the \emph{efficiency} of our programs. The possibility of using persistent homology techniques for the actual problem of fingerprint \emph{recognition} is doubtful (we discuss about this at the end of the paper).

The organization of the paper is as follows. After this introduction, in Section \ref{sec:prst-hmlg} we present the main classical ideas of persistent homology and our new generalization to the case of integer coefficients. Then, in Sections~\ref{sec:efhm} and~\ref{sec:efhm_and_dvf} we introduce respectively the effective homology method and its relation with the combinatorial technique of discrete vector field. These techniques are applied in Sections~\ref{sec:dvf_for_digital_images} and~\ref{sec:dvf-for-persistent-homology} for the computation of homology groups and persistent homology of digital images. Section \ref{sec:imple_and_examples} presents the implementation of our algorithms and some examples of application. The paper ends with a section of conclusions and further work.

\section{Persistent homology}
\label{sec:prst-hmlg}

\subsection{Preliminaries}

Let us begin by introducing some basic definitions and results about persistent homology. For details, see \cite{EH10}.

\begin{definition}
\label{defn:sc-filtration}
Let $K$ be a simplicial complex. A (finite) \emph{filtration} of $K$ is a nested sequence of subcomplexes \(K^i \subseteq K\) such that
$$\emptyset= K^{0} \subseteq K^1 \subseteq K^{2} \subseteq \cdots \subseteq K^{m}=K
$$
\end{definition}

For every $i \leq j$ we have an inclusion map on the canonically associated chain complexes $\inc^{i,j}: C(K^i) \hookrightarrow C(K^j)$ and therefore we can consider the induced homomorphisms $f^{i,j}_n : H_n(K^i) \rightarrow H_n(K^j)$, for each dimension $n$. The filtration produces then for each degree $n$ a sequence of homology groups connected by homomorphisms:

$$
0=H_n(K^0) \rightarrow H_n(K^1) \rightarrow \cdots \rightarrow H_n(K^m) =H_n(K)
$$


\begin{definition}
\label{defn:persistent-homology}
The $n$-th \emph{persistent homology groups} of $K$, denoted by $H^{i,j}_n(K)\equiv H^{i,j}_n$, are the images of the homomorphisms $f^{i,j}_n$:
$$ H^{i,j}_n = \im f^{i,j}_n, \mbox{ for } 0 \leq i \leq j \leq m$$

The group $H^{i,j}_n$ consists of the $n$-th homology classes of $K^i$ that are still alive at $K^j$. A class $\gamma \in H_n(K^i)$ is said to \emph{be born} at $K^i$ if $\gamma \notin H^{i-1,i}_n$. It is said to \emph{die} entering $K^j$ if it merges with an older class as we go from $K^{j-1}$ to $K^j$, that is, $f^{i,j-1}_n(\gamma) \notin H^{i-1,j-1}_n$ but $f^{i,j}_n(\gamma) \in H^{i-1,j}_n$. If $\gamma$ is born at $K^i$ and dies entering $K^j$, the difference $j-i$ is called the \emph{persistence index} of $\gamma$, denoted $\pers(\gamma)$. If $\gamma$ is born at $K^i$ but never dies then $\pers(\gamma)=\infty$.
\end{definition}

\subsection{Generalization to the integer case}

The main references on persistent homology \cite{EH10}, \cite{EH08} or \cite{ZC05} consider the case of coefficients over  a field. In that situation each group $H^{i,j}_n$ is a vector space which is determined up to isomorphism by its dimension, denoted $\beta^{i,j}_n$. This allows one to represent all persistent homology groups by means of a \emph{barcode} diagram~\cite{EH10}. However, if we work with $\Zset$-coefficients one can face extension problems. In order to solve this difficulty, in \cite{RHRS13} we introduced  the following generalization of persistent homology with $\Zset$-coefficients which leads to a new (more general) definition of barcode.

First of all one can observe that the groups $H^{\ast,j}_n$ provide a filtration of $H^j_n\equiv H_n(K^j)$:
\begin{equation*}
\label{eq:persistent-filtration}
0=H ^{0,j}_n \subseteq H^{1,j}_n \subseteq H^{2,j}_n \subseteq \cdots \subseteq H^{j,j}_n=H^j_n
\end{equation*}

We can then consider a \emph{double filtration} of $H^j_n$ obtained by introducing the \emph{new} groups $H^{i,j,k}_n$, for $i \leq j \leq k$, defined as
$$
H^{i,j,k}_n=H^{i,j}_n \cap (f_n^{j,k})^{-1}(H^{i-1,k}_n) \subseteq H^j_n
$$

For each fixed $i$ and $j$, the different groups $H^{i,j,\ast}_n$ define a filtration between $H^{i-1,j}_n$ and $H^{i,j}_n$:
\begin{equation*}
\label{eq:persistent-filtration-2}
H^{i-1,j}_n= H_n^{i,j,j} \subseteq H^{i,j,j+1}_n \subseteq  H^{i,j,j+2}_n \subseteq \cdots \subseteq H^{i,j,m}_n \subseteq  H^{i,j}_n
\end{equation*}

Each group $H^{i,j,k}_n$ contains all classes which are in $H^{i-1,j}_n$ and also the classes of $H^j_n$ which are born at $K^i$ and die at or before $K^k$.




Elementary  linear algebra  proves the  following quotient  groups are
canonically  isomorphic and  we denote  by $BD_n^{i,k}(K)$  their common
isomorphism class:

\begin{equation*}
\begin{aligned}
BD_n^{i,k}(K)  = & \frac{H_n^{i,i,k}}{H_n^{i,i,k-1}}  = \frac{H_n^{i,i+1,k}}{H_n^{i,i+1,k-1}} = \ldots \\
              & \ldots =\frac{H_n^{i,k-2,k}}{H_n^{i,k-2,k-1}}= \frac{H_n^{i,k-1,k}}{H_n^{i-1,k-1}}
\end{aligned}
\end{equation*}

 \noindent the notation  $BD_n^{i,k}(K)$ being read \emph{the  group of homological
classes \underline{b}orn  at time i and \underline{d}ying  at time k},
in  fact  a group  of  equivalence  classes  modulo inferior  homology
groups.

Each group $BD_n^{i,k}(K)$ admits a canonical divisor presentation:
\[
BD_n^{i,k}(K) \cong \Zset_{d_n^{i,k,1}} \oplus ... \oplus \Zset_{d_n^{i,k,p_{n,i,k}}}
\]
every $\Zset$-index $\in {0} \cup [2, ...]$ dividing the next one.

For every $1 \leq \ell \leq  p_{n,i,k}$, a \emph{bar} is produced, a bar
emph{labelled}   $\Zset_{d_n^{i,k,\ell}}$,   to    be   installed   in   the
n-dimensional barcode diagram between times  i and k.  This is clearly
the  canonical generalization to the \emph{universal integer  case} of
the standard barcodes  with coefficients in a field.  Cancelling the
``torsion'' bars leaves the ordinary barcodes.

Let us emphasize that when working over a field, the group $BD^{i,k}_n$ of \mbox{$n$-dimensional} classes that are born at $K^i$ and die entering $K^k$ is uniquely determined (up to isomorphism) by its rank, denoted $\mu^{i,k}_n$, which is given by the formula:
$$
\mu^{i,k}_n = (\beta^{i,k-1}_n - \beta^{i,k}_n) - (\beta_n^{i-1,k-1} - \beta_n^{i-1,k})
$$
so that the groups $BD^{i,k}_n$ can be determined if the groups $H^{i,j}_n$ are known. Conversely, in the field situation the information about the ranks of $BD^{i,k}_n$ is sufficient to know the \emph{total} groups $H^{i,j}_n$ and also the groups $H^{i,j,k}_n$ defined in our double filtration.

However, in the integer coefficient case the situation is not so favorable. Now the groups $H^{i,j}_n$ or $H^{i,j,k}_n$  are not sufficient to determine $BD^{i,k}_n$, because there could be several possibilities for the corresponding quotients\footnote{For example, the quotient of $\Zset_2 \oplus \Zset_4$ by $\Zset_2$ could be $\Zset_2 \oplus \Zset_2$ or $\Zset_4$.}. Similarly, from the groups $BD_n^{i,k}$ it is not always possible to determine the persistent homology groups $H^{i,j}_n$ and $H^{i,j,k}_n$ because of extension problems. For this reason, the classical representation of  persistent homology groups by means of a barcode diagram \cite{EH10} is not sufficient. See \cite{RHRS13} for more details about our definition of persistent homology and barcode diagram in the integer case.

\section{Effective homology}
\label{sec:efhm}

\subsection{Main definitions}

The effective homology method, introduced in \cite{Ser94} and explained in depth in~\cite{RS02} and~\cite{RS06}, is a technique which can be used to determine homology and homotopy groups of complicated spaces. We present now the main definitions and ideas of this method. All chain complexes considered in this section are chain complexes of free $\Zset$-modules.

\begin{definition}
\label{def:red}
A \emph{reduction} $\rho\equiv(D \rrdc C)$ between two
chain complexes $D$ and $C$ is a triple $(f,g,h)$ where: (a) The components
$f$ and $g$ are chain complex morphisms $f: D \rightarrow C$ and $g: C \rightarrow D$; (b)
The component $h$ is a homotopy operator $h:D\rightarrow D$ (a graded group homomorphism of degree +1); (c) The following relations are satisfied:
  (1) $f  g = \mbox{id}_C$; (2)
  $g f + d_D h + h  d_D
        = \mbox{id}_D$;
  (3)~\ {$f  h = 0$;} (4) $h   g = 0$; (5) $h   h = 0$.

\end{definition}

\begin{remark}
These relations express that $D$ is the direct sum of $C$ and a contractible (acyclic) complex. This decomposition is simply $D=\kr  f \oplus \im  g$, with $\im  g\cong C$ and $H_n(\kr  f )=0$ for all $n$. In particular, this implies that the graded homology groups \(H_\ast(D)\) and \(H_\ast(C)\) are canonically isomorphic.
\end{remark}

\begin{definition}
A \emph{(strong chain) equivalence} $\varepsilon \equiv (C \lrrdc E)$ between two complexes $C$ and~$E$ is a triple $(D,\rho,\rho ')$ where $D$ is a chain complex and $\rho$ and $\rho'$ are reductions from $D$ over $C$ and  $E$ respectively: $C \stackrel{\rho}{\lrdc} D \stackrel{\rho'}{\rrdc} E.$
\end{definition}

\begin{remark}
An effective chain complex is essentially a free chain complex $C$ where each group $C_n$ is finitely generated, and there is an algorithm that returns a $\Zset$-base in each degree $n$ (for details, see \cite{RS02}). The homology groups of an effective chain complex $C$ can easily be determined by means of some diagonalization algorithms on matrices (see \cite{KMM04}).
\end{remark}

\begin{definition}
An \emph{object with effective homology} is a triple $(X,EC,\varepsilon)$ where $EC$ is an effective chain complex and $\varepsilon$ is an equivalence between a free chain complex canonically associated to $X$ and $EC$, $C(X) \stackrel{\varepsilon}{\lrrdc} EC$.
\end{definition}

\begin{remark}
It is important to understand that in general the \(EC\) component of an object with effective homology is \emph{not} made of the homology groups of~\(X\); this component \(EC\) is a free \(\Zset\)-chain complex of finite type, in general with a non-null  differential, allowing to \emph{compute} the homology groups of \(X\); the justification is the equivalence \(\varepsilon\).
\end{remark}

The notion of object with effective homology makes it possible in this way to compute homology groups of complicated spaces by means of homology groups of effective complexes (which can easily be obtained using some elementary algorithms). The method is based on the following idea: given some topological spaces $X_1, \ldots, X_n$, a topological constructor $\Phi$ produces a new topological space $X$. If effective homology versions of the spaces $X_1, \ldots, X_n$ are known, then an effective homology version of the space $X$ can also be built, and this version allows us to compute the homology groups of $X$, even if it is not of finite type. A typical example of this kind of situation is the loop space
constructor. Given a {$1$-reduced} simplicial set $X$ with effective
homology, it is possible to determine the effective homology of
the loop space $\Omega(X)$, which in particular allows one to
compute the homology groups $H_\ast(\Omega(X))$. Moreover, if $X$
is \mbox{$m$-reduced}, this process may be iterated $m$ times, producing
an effective homology version of $\Omega^k(X)$, for $k \leq m$.
Effective homology versions are also
known for classifying spaces or total spaces of fibrations, see
\cite{RS06} for more information.

The effective homology method is implemented in a system called Kenzo~\cite{Kenzo}, a Lisp 16,000 lines program devoted to Symbolic Computation in Algebraic Topology, implemented by the third author of this paper and some coworkers. Kenzo works with rich and complicated algebraic structures (chain complexes, differential graded algebras, simplicial sets, simplicial groups, morphisms between these objects, reductions, etc.) and has obtained some results (for example homology groups of iterated loop spaces of a loop space modified by a cell attachment, components of complex Postnikov towers, homotopy groups of suspended classifying spaces, etc.) which had never been determined before. Kenzo has made it possible to detect an error in a theorem published in \cite{MW10}, where some theoretical reasonings are used to deduce that the fourth homotopy group of the suspended classifying space of the fourth alternating group $A_4$, $\pi_4(\Sigma K(A_4,1))$, is equal to $\Zset_4$; Kenzo's calculations have showed that the correct result (as later confirmed by the authors of \cite{MW10}) is $\Zset_{12}$. See \cite{RR12} for details on these calculations. Moreover, in \cite{RHRS13} Kenzo has been used to deduce the correct relation between persistent homology and spectral sequences and detect an error in \cite{EH10}: the so called ``Spectral sequence theorem'' \cite[p. 171]{EH10} includes a formula which is not correct (see \cite{RHRS13} for details).

\subsection{Relation with persistent homology}

The effective homology method makes it possible to compute homology groups of complicated spaces, even if they are not of finite type, by means of the notion of reduction. What about persistent homology? In this subsection we show that, if a reduction between two filtered chain complexes satisfies some ``natural'' conditions, then we will be able to determine the persistent homology groups of the big chain complex by means of those of the small one.

\begin{definition}
A (finite) \emph{filtration} of a chain complex $C$ is a family of sub-chain complexes $C^i \subseteq C$  such that
\[
0= C^0  \subseteq C^1 \subseteq C^2 \subseteq \cdots \subseteq C^m=C
\]
\end{definition}

\begin{definition}
Given two filtered chain complexes $C$ and $D$, a \emph{filtered chain complex morphism} $f: C \rightarrow D$ is a chain complex morphism which is compatible with the filtrations, that is to say,
$$
f(C^i_n) \subseteq D^i_n \mbox{ for each degree } n \mbox{ and filtration index } i.
$$
\end{definition}

\begin{definition}
Given two filtered chain complex morphisms $f,g: C \rightarrow D$ and a chain homotopy $h: f \simeq g$, we say that $h$ has \emph{order} $\leq s$ if
$$
h(C^i_n) \subseteq D^{i+s}_{n+1} \mbox{ for each degree } n \mbox{ and filtration index } i.
$$
\end{definition}

\begin{theorem} \cite{RHRS13}
\label{thm:prst-hom-efhm}
Let $C$ be a chain complex with a filtration. Let us suppose that $C$ is an object with effective homology, such that there exists an equivalence $ C \stackrel{\rho_1}{\lrdc} D \stackrel{\rho_2}{\rrdc} EC$ with $\rho_1=(f_1,g_1,h_1)$ and $\rho_2=(f_2,g_2,h_2)$, and such that filtrations are also defined on the chain complexes $D$ and $EC$. If the maps $f_1$, $f_2$, $g_1$, and $g_2$ are  morphisms of filtered chain complexes and both homotopies $h_1$ and $h_2$ have order $\leq s$, then the persistent homology groups $H^{i,j}_n$ of $C$ and $EC$ are (explicitly) isomorphic for $j-i\geq s$:
\[
H^{i,j}_n(C)\cong H^{i,j}_{n}(EC) \quad \mbox{for all } n \in \Nset \mbox{ and } j-i\geq s
\]
and the groups $BD^{i,k}_n$ of $C$ and $EC$ are (explicitly) isomorphic for $k-i > s$:
\[
BD^{i,k}_n(C)\cong BD^{i,k}_{n}(EC) \quad \mbox{for all } n \in \Nset \mbox{ and } k-i > s
\]
\end{theorem}

The isomorphisms between the corresponding groups are deduced from the compositions $f_2 g_1 : C \rightarrow EC$ and $f_1 g_2 : EC \rightarrow C$.

In particular, if both homotopies $h_1$ and $h_2$ have order $0$ (that is, they are compatible with the filtration on $D$), then all groups $H^{i,j}_n$ and $BD^{i,k}_n$ of $C$ and $EC$ are isomorphic.

Let us observe that if $EC$ is an effective chain complex, then one can determine its persistent homology groups by means of elementary algorithms: each subcomplex $EC^i$ has finite type, so that its homology groups $H_n(EC^i)\equiv H^i_n$ are computable. Then the maps $f^{i,j}_n:H^{i}_n \rightarrow H^{j}_n$ can be expressed by means of finite matrices and therefore we can compute the groups $H^{i,j}_n=\im f^{i,j}_n$. Similarly, the groups $H^{i,j,k}_n=H^{i,j}_n \cap (f_n^{j,k})^{-1}(H^{i-1,k}_n) \subseteq H^{i,j}_n \subseteq H^j_n$ and $BD_n^{i,k}={{H}^{i,i,k}_n}/{{H}^{i,i,k-1}_n}$ of $EC$ can be computed by means of matrix diagonalization. Thanks to Theorem \ref{thm:prst-hom-efhm}, we can also compute the persistent homology groups of the initial (big) chain complex $C$ by means of those of~$EC$.

Once a strong chain equivalence is established between an initial chain complex $C$ and an effective chain complex $EC$, we can obtain the generators of the homology groups of $C$ expressed as cycles on $C$. To this aim, we get the (representatives of) generators of the homology groups of $EC$ as cycles in $EC$ (it is a byproduct of the diagonalization process to determine Betti numbers and torsion coefficients), and then we apply on them the composition $f_1 g_2$ of the chain equivalence $\varepsilon$, getting the announced cycles over $C$. In fact, the chain equivalence produces a complete solution of the \emph{homological problem} for $C$; see the statement of this problem in \cite{Ser09}. If the chain complexes $C$ and $EC$ are filtered and the equivalence $\varepsilon$ satisfies the hypothesis of Theorem \ref{thm:prst-hom-efhm}, the same process as before produces the generators of the persistent homology groups $H^{i,j}_n(C)$ and $BD^{i,k}_n(C)$. This opens the possibility of a \emph{qualitative} study of persistent homology, going beyond the traditional \emph{quantitative} analysis (based, for instance, in barcodes). With our approach we can trace the born and death moments of particular cycles, and their contribution to the persistent homology groups.

The problem now is: given a filtered (big) chain complex $C$,  how can we obtain an equivalence $ C \stackrel{\rho_1}{\lrdc} D \stackrel{\rho_2}{\rrdc} EC$ where $EC$ is effective? A useful way to obtain such an equivalence (or more concretely, a reduction $\rho: C \rrdc EC$) consists in using \emph{discrete vector fields}, a combinatorial tool introduced in the following section.

\section{Effective homology and discrete vector fields}
\label{sec:efhm_and_dvf}

The notion of discrete vector field (DVF) is due to Robin Forman \cite{For98}; it is an
essential component of the so-called discrete Morse theory.
This notion is usually described and used in combinatorial topology, but a
purely algebraic version can also be given as follows. See \cite{RS10} for more details.

\begin{definition}
\label{defn:dvf}
Let $C=(C_n,d_n)_{n \in \Zset}$ be a free chain complex with distinguished $\Zset$-basis $\beta_n \subset C_n$. A \emph {discrete vector field} $V$ on $C$ is a collection of pairs $V = \{(\sigma_i; \tau_i)\}_{i\in I}$ satisfying the conditions:
\begin{itemize}
\item Every $\sigma_i$ is some element of some $\beta_n$, in which case $\tau_i\in \beta_{n+1}$.
The degree $n$ depends on $i$ and in general is not constant.
\item Every component $\sigma_i$ is a \emph{regular face} of the corresponding $\tau_i$ (that is, the coefficient of $\sigma_i$ in $d\tau_i$ is $+1$ or $-1$).
\item Each generator (\emph{cell}) of $C$ appears at most one time in $V$.
\end{itemize}
\end{definition}

It is not required all the cells of $C$ appear in the vector field $V$. In particular the
void vector field is allowed. In a sense the remaining cells are the most important. Moreover, we do not assume the distinguished bases $\beta_n$ are finite, the chain groups $C_n$
are not necessarily of finite type.

\begin{definition}
A cell $\sigma\in \beta_n$ which does not appear in the discrete vector field $V$ is called a \emph{critical cell}.
\end{definition}

In the case of a chain complex coming from a topological cellular complex, a DVF is
a recipe to cancel ``useless'' cells in the underlying space, useless with
respect to the homotopy type. For example \(\partial
\Delta^2\) and the circle have the same homotopy type, which is described by
the following scheme: \vspace{-10pt}
\begin{equation*}\begin{tikzpicture} [scale = 0.5, baseline = (baseline)]
 \coordinate (baseline) at (0,0.5) ;
 \coordinate (0) at (0,0) ;
 \coordinate (1) at (1,2) ;
 \coordinate (2) at (2,0) ;
 \begin{scope} [font = \footnotesize]
 \node [anchor = 0] at (0) {0} ;
 \node [anchor = 0] at (1) {1} ;
 \node [anchor = 180] at (2) {2} ;
 \node [anchor = 0] at (6,0) {0} ;
 \node [anchor = 0] at (0.5,1) {01} ;
 \node [anchor = 180] at (1.5,1) {12} ;
 \node [anchor = 90] at (1,0) {02} ;
 \draw (6,0) .. controls +(60:4) and +(0:4) .. node [right] {12} (6,0) ;
 \end{scope}
 \foreach \i in {0,1,2} {\node at (\i) {\(\bullet\)} ;}
 \draw (0) \foreach \i in {1,2,0} {-- (\i)} ;
 \draw [thick, ->] (1) -- (0.5,1) ;
 \draw [thick, ->] (2) -- (1,0) ;
 \node at (4,1) {\(\Longrightarrow\)} ;
 \node at (6,0) {\(\bullet\)} ;
\end{tikzpicture}
\end{equation*}

The initial simplicial complex is made of three 0-cells 0, 1 and 2, and three
1-cells 01, 02 and 12. The drawn vector field is \(V = \{(1;01), (2;02)\}\),
and this DVF defines a homotopy equivalence between
\(\partial\Delta^2\) and the minimal triangulation of the circle as a
simplicial set.

\begin{definition}
Given a discrete vector field $V$, a \emph{$V$-path $\pi$ of
degree $n$ and length $m$} is a sequence $\pi =\{(\sigma_{i_k}; \tau_{i_k})\}_{0 \leq k <
m}$ satisfying:
\begin{itemize}
\item
Every pair $(\sigma_{i_k}; \tau_{i_k})$ is a component of
$V$ and $\tau_{i_k}$ is an $n$-cell.
\item
For every $0 < k < m$, the component $\sigma_{i_k}$ is a face of
$\tau_{i_{k-1}}$, non necessarily regular, but different from
$\sigma_{i_{k-1}}$.
\end{itemize}
The $V$-path is said \emph{starting from} $\sigma_{i_0}$.
\end{definition}

\begin{definition}
A discrete vector field $V$ is \emph{admissible} if for
every $n \in \Zset$, a function $\lambda_n: \beta_n \rightarrow \Nset$ is
provided satisfying the following property: every $V$-path starting from
$\sigma \in \beta_n$ has a length bounded by $\lambda_n(\sigma)$.
\end{definition}

The admissability property is necessary to exclude infinite paths and loops, and is used to ensure the homotopy type of the corresponding \emph{reduced} chain complex (see Theorem \ref{thm:dfv-reduction}) is the same as the initial one.

Discrete vector fields and effective homology can be related as follows. Let $C=(C_n,d_n)_{n \in \Zset}$  be a free chain complex provided with an admissible
discrete vector field $V$. Then a reduction $\rho : C \rrdc C^c$
can be constructed where the small chain complex $C^c$
is the \emph{critical complex}; it is also a cellular complex but
generated only by the critical cells of $C$, those which do not appear in the vector
field $V$, with a differential appropriately defined (combining the initial differential $d$
of $C$ and the DVF). This result is due to Robin Forman \cite[Section 8]{For98}, and can be extended to
complexes not necessarily of finite type. In \cite{RS10} two different proofs of this result are given.

\begin{theorem}\cite{RS10}[Vector-Field Reduction Theorem]
\label{thm:dfv-reduction}
Let $C=(C_n,d_n)_{n \in \Zset}$ be
a free chain complex and $V = \{(\sigma_i; \tau_i)\}_{i\in I}$ be an admissible discrete vector
field on $C$. Then the vector field $V$ defines a canonical reduction $\rho = (f, g, h) :
(C_n, d_n)\rrdc (C^c_n, d'_n)$ where $C^c_n = \Zset[\beta^c_n]$ is the free $\Zset$-module generated by the critical
$n$-cells and $d'_n$ is an appropriate differential canonically deduced from $C$ and $V$.
\end{theorem}

\textbf{Proof}\\
A cell basis \(\beta_n\) is canonically divided by the vector field \(V\) into
three components \(\beta_n = \beta^t_n + \beta^s_n + \beta^c_n\) where
\(\beta^t_n\) (resp. \(\beta^s_n\), \(\beta^c_n\)) is made of the target (resp.
source, critical) cells. For the third condition of
Definition~\ref{defn:dvf} implies a cell cannot be simultaneously a source cell
and a target cell.

The decompositions of the bases \(\beta_n\) induce a
corresponding decomposition of the chain groups \(C_n = C^t_n \oplus C^s_n
\oplus C^c_n\), so that every differential \(d_n\) can be viewed as a \(3
\times 3\) matrix

{\footnotesize\begin{equation*} d_n =
\left[\begin{array}{ccccc}
  d_{n,1,1} & d_{n,1,2} & d_{n,1,3}
  \\
  d_{n,2,1} & d_{n,2,2} & d_{n,2,3}
  \\
  d_{n,3,1} & d_{n,3,2} & d_{n,3,3}

\end{array} \right]
\end{equation*}}

It can be seen that the component $d_{n,2,1}: C_n^t \rightarrow C_{n-1}^s$ is an isomorphism and its inverse map $d^{-1}_{n,2,1} : C_{n-1}^s \rightarrow C_n^t$ can be given by the recursive formula:
\begin{equation*}
d_{n,2,1}^{-1}(\sigma) = \varepsilon(\sigma, V(\sigma))\,\left(V(\sigma) -
\sum_{\sigma' \in \beta^s_{n-1} - \{\sigma\}} \varepsilon(\sigma', V(\sigma))
\,d_{n,2,1}^{-1}(\sigma')\right)
\end{equation*}
where the \emph{incidence number} \(\varepsilon(\sigma, \tau)\) is the
coefficient of \(\sigma\) in the differential \(d\tau\). Let us remark that
\(\varepsilon(\sigma, V(\sigma)) = \pm 1\), but the other incidence numbers
are arbitrary.

The maps $d'$, $f$, $g$ and $h$  are then given by the formulas:
\begin{equation*}
 \begin{array}{c@{\hspace{20pt}}c@{\hspace{20pt}}}
 d'_n = d_{n,3,3} - d_{n,3,1} d_{n,2,1}^{-1} d_{n,2,3}
 &
 f_n = \left[\begin{array}{ccc} 0 & - d_{n,3,1} d_{n,2,1}^{-1} & 1 \end{array}\right]
 \\[10pt]
 g_n = \left[\begin{array}{c} - d_{n,2,1}^{-1} d_{n,2,3} \\ 0 \\ 1
 \end{array}\right]
 &
 h_n = \left[\begin{array}{ccc} 0 & d_{n,2,1}^{-1} & 0 \\ 0 & 0 & 0
 \\ 0 & 0 & 0 \end{array}\right]
 \end{array}
\end{equation*}

In this way, the maps $d'$, $f$, $g$ and $h$  which provide the reduction $\rho$ can be explicitly constructed.

Thanks to Theorem \ref{thm:dfv-reduction} and the explicit isomorphism $H_n(C) \cong H_n(C^c)$ deduced from the reduction $\rho: C \rrdc C^c$, one can compute the homology groups of the big (possibly infinite) chain complex $C$ by means of the homology groups of $C^c$ (whenever $C^c$ is of finite type). In \cite{RS10}, this result is used to produce, for example, the effective homology of twisted cartesian products and Eilenberg-MacLane spaces.

Let us present now the first new result of this paper, where we extend the result of Theorem \ref{thm:dfv-reduction} for the computation of persistent homology groups.

 \begin{definition}
Let $C=(C_n,d_n,\beta_n)_{n \in \Zset}$ be a free chain complex provided with a filtration $0= C^0  \subseteq C^1 \subseteq C^2 \subseteq \cdots \subseteq C^m=C$. We say that a generator $\sigma \in \beta_n$ has \emph{filtration index} $i$ if $\sigma \in C^i_n$ and $\sigma \notin C^{i-1}_n$.
\end{definition}

\begin{theorem}
\label{thm:dvf-persistenthomology}
Let $C=(C_n,d_n,\beta_n)_{n \in \Zset}$ be
a free chain complex with a filtration and $V = \{(\sigma_i; \tau_i)\}_{i\in I}$ be an admissible discrete vector
field on $C$ such that for every $i \in I$ the elements $\sigma_i$ and $\tau_i$ have the same filtration index. Then the canonical reduction $\rho = (f, g, h) :
(C_n, d_n)\rrdc (C^c_n, d'_n)$ described in Theorem \ref{thm:dfv-reduction} is compatible with the filtration. In particular, the chain homotopy $h$ has necessarily order $\leq 0$.
\end{theorem}

\textbf{Proof} \\
As seen in the proof of Theorem \ref{thm:dfv-reduction}, the differential map $d'$ of the critical chain complex $C^c$ is defined by means of the initial differential $d$ and the discrete vector field $V$. In particular, one can observe that if every vector in the discrete vector field is made of two elements with the same \emph{filtration index} then the new differential $d'$ is compatible with the filtration and therefore it induces in a natural way a filtration on the critical chain complex $C^c$.

Similarly, the maps $f$, $g$ and $h$ in the reduction $\rho$ are also obtained as certain combinations of $d$ and the DVF. As before, one can easily deduce from the corresponding formulas that if every vector in the discrete vector field is made of two elements with the same filtration index, then the morphisms $f$, $g$ and $h$ will be compatible with the filtration.

\begin{corollary}
\label{cor:dvf-persistenthomology}
Let $C=(C_n,d_n,\beta_n)_{n \in \Zset}$ be
a free chain complex with a filtration and $V = \{(\sigma_i; \tau_i)\}_{i\in I}$ be an admissible discrete vector
field on $C$ such that for every $i \in I$ the elements $\sigma_i$ and $\tau_i$ have the same filtration index. Then
all persistent homology groups $H^{i,j}_n$ and $BD^{i,k}_n$ of the critical complex $C^c$ and those of $C$ are (explicitly) isomorphic.
\end{corollary}

\textbf{Proof} \\
We apply Theorem \ref{thm:prst-hom-efhm} to the reduction $\rho = (f, g, h) :
(C_n, d_n)\rrdc (C^c_n, d'_n)$. It is a particular case of strong chain equivalence, where  the left reduction $\rho_1$ is trivial ($f_1$ and $g_1$ are the corresponding identity maps and $h_1$ is null), and $\rho_2=\rho$.

Our new results Theorem \ref{thm:dvf-persistenthomology} and Corollary \ref{cor:dvf-persistenthomology} allow us to compute persistent homology groups of big chain complexes by means of those of a smaller chain complex obtained from a DVF. Let us see in the following section how we can determine a DVF for the particular case of digital images. 

\section{Discrete vector fields for digital images}
\label{sec:dvf_for_digital_images}

In this section we present an algorithm producing DVFs for digital images. The algorithm is included in the unpublished paper~\cite{RS10} and in~\cite{PDHR13} has been formally verified by means of the interactive theorem prover Coq~\cite{Coq}.

Since we have placed ourselves in an algebraic setting, we identify a digital image with its algebraic counterpart, as explained in the following definition.

\begin{definition}
\label{defn:digital_image}
A \emph{digital image}, an \emph{image} in short, is a \emph{finite}
algebraic cellular complex \((C_n, d_n, \beta_n)_{n \in \Zset}\): every
\(\beta_n\) is finite and furthermore every \(\beta_n\) is empty outside an
interval \([0\ldots N]\), the smallest possible \(N\) being the
\emph{dimension} of the image.
\end{definition}

For example the various techniques of scientific imaging produce \emph{images},
some finite objects, typically finite sets of pixels. If you are interested in
some \emph{homological} analysis of such an image, you associate to it a
\emph{geometrical} cellular complex, again various techniques can be used, and
finally this defines an algebraic cellular complex, the complex defining the homology groups of the geometrical object. There remains to compute
the homology groups of this complex; in fact computing the \emph{effective}
homology of this complex is much better.

The Vector-Field Reduction Theorem (Theorem~\ref{thm:dfv-reduction}) is then particularly
welcome. The bases \(\beta_n\) of the initial complex can be enormous, but
appropriately choosing a vector field can produce, applying this Reduction
Theorem, a new complex which is homology equivalent, with small \emph{critical}
bases \(\beta^c_n\); so that the homology computations are then fast.

We begin by considering the simplest case of a chain complex with only two consecutive chain groups, and then the general case is easily reduced to this one.



Let \(M\) be a matrix \(M \in \textrm{Mat}_{m,n}(\Zset)\), with \(m\)
rows and \(n\) columns. Think of \(M\) as the unique non-null boundary matrix
of the chain complex:
\begin{equation*}
C: \cdots \leftarrow 0 \leftarrow \Zset^m \stackrel{M}{\longleftarrow} \Zset^n
\leftarrow 0 \leftarrow \cdots
\end{equation*}

A \emph{vector field} \(V\) for this matrix is nothing but a set of integer
pairs \(\{(a_i; b_i)\}_i\) satisfying these conditions:
\begin{enumerate}
\item
\(1 \leq a_i \leq m\) and \(1 \leq b_i \leq n\).
\item
The entry \(M[a_i, b_i]\) of the matrix is \(\pm 1\).
\item
The indices \(a_i\) (resp. \(b_i\)) are pairwise different.
\end{enumerate}

This clearly corresponds to a DVF for the chain complex $C$, and constructing such a vector
field is very easy. But there remains the main problem: is this vector
field \emph{admissible}? Because the context is finite, it is a matter of avoiding \emph{loops}. If
the vector field is admissible, it defines a \emph{partial} order between
source cells: the relation \(a > a'\) is satisfied between source cells if and
only if a \mbox{\(V\)-path} goes from \(a\) to \(a'\). The non-existence of loops
guarantees this is actually a partial order.

Conversely, let \(V\) be a vector field for our matrix \(M\). If \(1 \leq a, a'
\leq m\), with \(a \neq a'\), we can decide \(a > a'\) if there is an
\emph{elementary} V-path from \(a\) to \(a'\), that is, if a vector \((a;b)\)
is present in \(V\) and the entry \(M[a', b]\) is non-null; for this
corresponds to a cell \(b\) with in particular \(a\) as regular face and \(a'\)
as an arbitrary face. We so obtain a binary relation. Then the vector field
\(V\) is admissible if and only if this binary relation actually transitively
generates a partial order, that is, if again there is no loop \(a_1 > a_2 >
\cdots > a_k = a_1\).

\begin{definition}
\label{defn:maximal_dvf}
Let $V$ be an admissible discrete vector field on a chain complex~$C$. $V$ is said to be \emph{maximal} if it is not possible to add a new vector $(\sigma;\tau)$ to $V$ such that the new vector field $V':=V \cup \{(\sigma;\tau)\}$ is admissible.
\end{definition}

Let us remark that a DVF being maximal does not imply its number of vectors is maximal. Finding a vector field of maximal
size seems much too difficult in real applications. Finding a maximal admissible
vector field, not the same problem, is more reasonable but still serious.

{A direct way to quickly construct an admissible DVF for a matrix consists in
\emph{pre}defining an order between row indices, and to collect all the indices
for which some column is ``above this index''. Let us play with this toy-matrix
given by our random generator:}\vspace{-10pt}

 {\footnotesize\begin{equation*} M =
\left[\begin{array}{ccccc}
  0 & 0 & -1 & -1 & 0
  \\
  0 & -1 & 0 & 0 & 1
  \\
  0 & 0 & 0 & 1 & 1
  \\
  0 & -1 & 1 & 0 & -1
  \\
  -1 & 1 & -1 & 0 & 0
\end{array} \right]
\end{equation*}}

If we take simply the index order between row indices, we see the columns 1, 4
and 5 can be selected, giving the vector field \(\{(5;1), (3;4), (4;5)\}\).
Theorem~\ref{thm:dfv-reduction} produces a reduction of \(M: \Zset^5 \leftarrow \Zset^5\) to
\(M': \Zset^2 \leftarrow \Zset^2\) with

{\footnotesize\begin{equation*} M' = \left[\begin{array}{cc}
 -1 & 0 \\ -2 & 1
\end{array}\right]
\end{equation*}}%

In other words, a reduction is constructed from the initial chain complex $\cdots \leftarrow 0 \leftarrow \Zset^5 \stackrel{M}{\longleftarrow} \Zset^5
\leftarrow 0 \leftarrow \cdots$ to a smaller chain complex $\cdots \leftarrow 0 \leftarrow \Zset^2 \stackrel{M'}{\longleftarrow} \Zset^2
\leftarrow 0 \leftarrow \cdots$. See \cite{RS10} for details on the explicit formula for $M'$.

A more sophisticated strategy consists, given an \emph{admissible} vector field
already constructed, in trying to add a new vector to obtain a better
reduction. The already available vector field defines a partial order between
the source cells with respect to this vector field and the game now is to
search a new vector to be added, but keeping the admissibility property. This
process is applied by  starting from the void DVF.

Let us try to apply this process to the same matrix $M$ as before.
We start with the void vector field \(V_0 = \{\}\). Running the successive rows
in the usual reading order, we find \(M[1,3] = -1\), and we add the vector
(1;3), obtaining \(V_1 = \{(1;3)\}\). Only one source cell 1, but we must note
that it is from now on forbidden to add a vector which would produce the
relation \(4 > 1\) or \(5 > 1\): this will generate a loop \(1 > 4 > 1 > \cdots
\) and the same for 5. In other words, the partial order to be recorded is \(1
> 4\) and \(1 > 5\), even if 4 and 5 are not yet source cells. Also the row 1
and the column 3 are now used and cannot be used anymore.

We read the row 2 and find \(M[2,2] = -1\), which suggests to add the vector
\((2;2)\), possible, with the same restrictions as before. Now \(V_2 = \{(1;3),
(2;2)\}\).

Reading the row 3 suggests to add the vector \((3;4)\) where 4 has 1 as a face,
because \(M[1,4] = -1\). This does not create any cycle, and we define \(V_3 =
\{(1;3), (2;2), (3;4)\}\). We note also that \(3 > 1\).

Reading the row 4, the only possibility would be the new vector \((4;5)\), but
2 is a face of 5 and this would generate the loop \(4 > 2 > 4 > \cdots\),
forbidden. It is impossible to add a vector \((4;-)\).

Finally we can add the vector \((5;1)\), convenient, for 1 has no other face
than~5; adding this vector certainly keeps the admissibility property.

This leads to the maximal vector field \(V_4 = \{(1;3), (2;2), (3;4), (5;1)\}\), which
generates the partial order on \(1,2,3,5\) where the only non-trivial relations
are \(3 > 1 > 5\) and \(2 > 5\). 
Reordering the rows and columns in the respective
orders \((3,1,2,5,4)\) and \((4,3,2,1,5)\) gives the new form for our
matrix:

{\footnotesize\begin{equation*}
 M = \left[\begin{array}{ccccc}
 1 & 0 & 0 & 0 & 1
 \\
 -1 & -1 & 0 & 0 & 0
 \\
 0 & 0 & -1 & 0 & 1
 \\
 0 & -1 & 1 & -1 & 0
 \\
 0 & 1 & -1 & 0 & -1
 \end{array}\right]
\end{equation*}}

The DVF has 4 components, and the \(4 \times 4\) top left-hand
submatrix is triangular unimodular. The reduction produces the matrix
\(\left[-1\right]\) which of course can be reduced to the void matrix.

The example shows a first step of reduction produces a smaller matrix which in
turn can sometimes be also reduced, even if the used vector field is maximal.

We obtain in this way an algorithm producing a maximal admissible DVF which allows one to reduce the number of generators of the initial chain complex.

\begin{theorem}
\label{thm:matrix-dvf}
An algorithm can be written down:
\begin{itemize}
\item \textbf{Input:} A matrix $M\in \textrm{Mat}_{m,n}(\Zset)$.
\item \textbf{Output:} A maximal admissible discrete vector field $V$ for $M$.
\end{itemize}
\end{theorem}

See \cite{RS10} for more details on the algorithm and some considerations about a graph interpretation which can be useful for more realistic (that is, for bigger matrices) situations.

Let $C$ be a chain complex of finite type with only two non-null consecutive chain groups $\cdots \leftarrow 0 \leftarrow \Zset^m \stackrel{M}{\longleftarrow} \Zset^n
\leftarrow 0 \leftarrow \cdots$. Applying Theorem \ref{thm:matrix-dvf} to the differential map matrix $M$, we obtain a maximal admissible discrete vector field $V$ for $C$. Considering now Theorem \ref{thm:dfv-reduction}, we obtain a reduction from the initial chain complex $C$ to a smaller (also effective) one $C^c =: EC$. In particular this allows one to compute the homology groups of the big chain complex $C$ by means of those of $EC=C^c$ in a more efficient way.

Let us consider now a general \emph{digital image} $C$ (see Definition \ref{defn:digital_image}). The first differential map $d_1: C_1 \rightarrow C_0$ is given by a matrix $M_1 \in \textrm{Mat}_{n_0,n_1}(\Zset)$. Applying Theorem \ref{thm:matrix-dvf}, we obtain a maximal admissible discrete vector field for $C$ and then Theorem \ref{thm:dfv-reduction} produces a reduction to a smaller chain complex $C^c =: EC^1$. We consider now the second differential map of $EC^1$, $d_2: EC^1_2 \rightarrow EC^1_1$ given by a matrix $M^1_2$, and apply again Theorems \ref{thm:matrix-dvf} and \ref{thm:dfv-reduction}. This produces a new reduction from $EC^1$ to a new smaller chain complex $(EC^1)^c =: EC^2$. The process can be iterated for every dimensions $n\leq N$. The compositions of all reductions produces a reduction from the initial chain complex $C$ to the \emph{last} chain complex $EC^N \equiv EC$. This leads to the following theorem.

\begin{theorem}
\label{thm:image-dvf}
An algorithm can be written down:
\begin{itemize}
\item \textbf{Input:} A digital image $C$.
\item \textbf{Output:} A reduction $\rho: C \rrdc EC$ obtained as a composition of several reductions, each of them deduced from a maximal admissible discrete vector field applying Theorem \ref{thm:dfv-reduction}.
\end{itemize}
\end{theorem}

The fact that each DVF involved in the construction of $\rho$ is maximal implies that the chain complex $EC$ is usually significantly smaller than $C$. The homology groups of $C$ can then be determined in a more efficient way.

Let us consider the following toy
example, a screen with a \(3 \times 3\) ``resolution'' and this image, eight
pixels black and one white.
\begin{equation*}\begin{tikzpicture} [scale = 0.5, baseline = 0.7cm]
 \fill [black!30] (0,0) rectangle (3,3) ;
 \fill [white] (1,1) rectangle (2,2) ;
 \foreach \i in {0,...,3}
   {\draw (\i, 0) -- +(0,3) ;
    \draw (0, \i) -- +(3,0) ;
   }
\end{tikzpicture}\end{equation*}

The bases of the corresponding cellular complex $C$ are made of 16 vertices, 24
edges and 8 squares. Theorem \ref{thm:image-dvf} could produce the following vector field, with only two critical cells (one vertex and one edge).
\begin{equation*}\begin{tikzpicture} [scale = 0.5, baseline = 0.7cm]
 \fill [black!30] (0,0) rectangle (3,3) ;
 \fill [white] (1,1) rectangle (2,2) ;
 \foreach \i in {0,...,3}
   {\draw (\i, 0) -- +(0,3) ;
    \draw (0, \i) -- +(3,0) ;
   }
 \foreach \i in {0,0.5,...,3}
   \foreach \j in {1,3}
     {\draw [->, thick] (\i,\j) node {\scriptsize\(\bullet\)} -- +(0,-0.5) ;}
 \foreach \i in {0,0.5,1,2,2.5,3}
   {\draw [->, thick] (\i,2) node {\scriptsize\(\bullet\)} -- +(0,-0.5) ;}
 \foreach \i in {1,2,3}
   {\draw [->, thick] (\i,0) node {\scriptsize\(\bullet\)} -- +(-0.5,0) ;}
\end{tikzpicture}\end{equation*}

The reduced chain complex is \(0 \leftarrow \Zset \leftarrow \Zset \leftarrow 0\)
with the null map between both copies of \(\Zset\). The reduction described by Theorem \ref{thm:dfv-reduction} informs that \(H_0(C) = H_1(C)=\Zset\) and produces a representant
for the generating homology classes.

\section{Discrete vector fields for computing effective persistent homology of digital images}
\label{sec:dvf-for-persistent-homology}

Let us suppose now that we have a digital image which is \emph{filtered}, that is, we consider that each pixel appears at some moment (filtration index) $i$. This gives us a list of images such that all pixels of each image are present in the following one. The associated chain complex is filtered and then it makes sense to compute its persistent homology groups, which can be used to determine some relevant features of the image in contrast with the noise. We can slightly modify the algorithm of Theorem \ref{thm:image-dvf} as follows such that the small chain complex $EC$ obtained by means of the construction of a DVF in each dimension allows us to compute the persistent homology groups of the initial image.

Let $C$ be a digital image with a filtration. The set of generators in each dimension $n$, $\beta_n$, can be reordered by their filtration index in an increasing way. With this order, each differential matrix $M_n: C_n \rightarrow C_{n-1}$ can be decomposed in blocks such that each submatrix in the diagonal corresponds to elements of the same filtration index. We begin by considering the first matrix $M_1: C_1 \rightarrow C_0$, with the previous decomposition. We apply Theorem \ref{thm:matrix-dvf} separately to each submatrix in the diagonal, which produces  several admissible discrete vector fields where both elements in each vector have the same filtration index. The vector fields are disjoint and it is possible to concatenate all of them, obtaining a discrete vector field which is therefore compatible with the filtration (let us note that this opens the possibility of \emph{parallel} processing, since the computation in each diagonal block is independent from the rest). The DVF $V$ so obtained is \emph{maximal with respect to the filtration}, that is, it is not possible to add a new vector $(\sigma;\tau)$ such that the new vector field $V':=V \cup \{(\sigma;\tau)\}$ is admissible and compatible with the filtration. Applying now Theorem \ref{thm:dvf-persistenthomology} and Corollary \ref{cor:dvf-persistenthomology}, we obtain a reduction from $C$ to a new filtered chain complex $C^c =: EC^1$ where the persistent homology groups of the reduced chain complex are isomorphic to those of the initial one. We repeat the procedure with the second differential map matrix of $EC^1$, $M^1_2: EC^1_2 \rightarrow EC^1_1$, obtaining a new reduction compatible with the filtrations. Iterating the process and composing the different reductions, we obtain a reduction from the initial filtered digital image $C$ to a smaller filtered chain complex $EC$, which allows one to determine the persistent homology of $C$ in an efficient way. This leads to the following theorem.

\begin{theorem}
\label{thm:image-dvf-persistenthomology}
An algorithm can be written down:
\begin{itemize}
\item \textbf{Input:} A filtered digital image $C$.
\item \textbf{Output:} A reduction $\rho: C \rrdc EC$ compatible with the filtration, obtained as a composition of several reductions, each of them constructed by applying Theorem \ref{thm:dfv-reduction} to an admissible discrete vector field which is maximal with respect to the filtration.
\end{itemize}
\end{theorem}

As explained for Theorem \ref{thm:image-dvf}, the fact that each DVF involved in the construction of $\rho$ is maximal with respect to the filtration implies that the chain complex $EC$ is usually significantly smaller than $C$. In this way, this result allows one to compute persistent homology of (big) digital images by means of those of a small chain complex. The technique can be considered as a generalization of~\cite{MN13}.

\section{Implementation and examples}
\label{sec:imple_and_examples}

The algorithm presented in Theorem \ref{thm:image-dvf-persistenthomology} has been implemented in a new module for the Kenzo system making use of programs previously developed. More concretely, in \cite{RHRS13} a new module for the Kenzo system was constructed allowing the computation of persistent homology of filtered chain complexes, making use of \emph{spectral sequences} and the effective homology technique. On the other hand, discrete vector fields were also previously implemented in Kenzo, including the construction of the associated reduction described in Theorem~\ref{thm:dfv-reduction}. Moreover, we have also made use of a new module for the Kenzo system for the computation of homology groups of digital images (allowing in particular the construction of the simplicial complex  associated with a digital image) developed by J\'onathan Heras \cite{Her11,HPR11}.

Our new module for Kenzo includes the implementation of algorithms described in Theorems \ref{thm:matrix-dvf}, \ref{thm:image-dvf} and \ref{thm:image-dvf-persistenthomology} and makes it possible to compute persistent homology groups of a filtered digital image $C$. First of all, the program reorders the generators of the image by increasing filtration index. Then it considers the first differential map $d_1$ and applies Theorem \ref{thm:matrix-dvf} on the different submatrices on the diagonal corresponding to elements of the same filtration index, producing a DVF on $C$. Then it computes by means of Theorem \ref{thm:dfv-reduction} the associated reduction to the critical complex $C^c \equiv EC^1$, repeats the process for the second differential map $d^1_2$ of $EC^1$, and so on. We obtain in this way the reduction described by Theorem \ref{thm:image-dvf-persistenthomology}, which is compatible with the filtration. This reduction is used as the effective homology of the initial image, and thanks to Theorem \ref{thm:prst-hom-efhm} we can compute the persistent homology groups of $C$ in an efficient way.

\begin{figure}
\begin{equation*}
\begin{tikzpicture} [scale = 0.15, baseline = 0.7cm]
 \fill [black!50] (0,15) rectangle (3,11) ;
 \fill [black!50] (3,11) rectangle (4,12) ;
     \fill [black!50] (9,15) rectangle (11,14) ;
     \fill [black!50] (9,12) rectangle (15,11) ;
       \fill [black!50] (14,15) rectangle (15,12) ;
       \fill [black!50] (4,11) rectangle (5,9) ;
    \fill [white] (14,15) rectangle (15,13) ;
    \fill [white] (12,12) rectangle (13,11) ;
    \fill [white] (1,14) rectangle (3,13) ;
 \fill [white] (1,11) rectangle (2,13) ;
      \fill [black!50] (7,10) rectangle (9,9) ;
    \fill [black!50] (0,10) rectangle (2,9) ;
    \fill [black!50] (1,9) rectangle (2,7) ;
    \fill [black!50] (3,6) rectangle (5,5) ;
    \fill [black!50] (4,9) rectangle (5,6) ;
    \fill [black!50] (0,4) rectangle (1,2) ;
    \fill [black!50] (1,3) rectangle (2,2) ;
    \fill [black!50] (3,2) rectangle (5,1) ;
    \fill [black!50] (11,11) rectangle (15,7) ;
    \fill [white] (12,11) rectangle (13,8) ;
    \fill [white] (14,10) rectangle (15,9) ;
    \fill [white] (12,12) rectangle (14,9) ;
    \fill [black!50] (8,8) rectangle (9,3) ;
    \fill [black!50] (9,7) rectangle (13,4) ;
    \fill [white] (8,6) rectangle (12,5) ;
    \fill [white] (10,5) rectangle (11,6) ;
    \fill [black!50] (11,6) rectangle (15,5) ;
    \fill [black!50] (10,3) rectangle (13,0) ;
    \fill [black!50] (13,3) rectangle (15,2) ;
    \fill [black!50] (14,2) rectangle (15,1) ;
    \fill [white] (11,2) rectangle (12,1) ;
    \fill [white] (10,0) rectangle (12,1) ;
    \fill [white] (11,7) rectangle (13,6) ;
    \fill [white] (8,4) rectangle (9,3) ;
    \fill [white] (10,4) rectangle (11,5) ;
    \fill [white] (11,5) rectangle (12,6) ;
    \fill [white] (7,10) rectangle (9,9) ;
    \fill [black!50] (13,12) rectangle (14,11) ;
 \foreach \i in {0,...,15}
   {\draw (\i, 0) -- +(0,15) ;
    \draw (0, \i) -- +(15,0) ;
   }
\end{tikzpicture}
\hspace{20pt}
\begin{tikzpicture} [scale = 0.15, baseline = 0.7cm]
 \fill [black!50] (0,15) rectangle (3,11) ;
 \fill [white] (1,14) rectangle (3,13) ;
 \fill [white] (1,11) rectangle (2,12) ;
 \fill [black!50] (3,11) rectangle (4,12) ;
     \fill [black!50] (9,15) rectangle (11,14) ;
     \fill [black!50] (9,12) rectangle (15,11) ;
    \fill [white] (12,12) rectangle (13,11) ;
    \fill [black!50] (14,15) rectangle (15,12) ;
    \fill [white] (14,15) rectangle (15,13) ;
    \fill [black!50] (3,11) rectangle (5,9) ;
    \fill [black!50] (7,10) rectangle (9,9) ;
    \fill [black!50] (0,10) rectangle (2,9) ;
    \fill [black!50] (1,9) rectangle (2,7) ;
    \fill [black!50] (3,6) rectangle (5,5) ;
    \fill [black!50] (4,9) rectangle (5,6) ;
    \fill [black!50] (0,4) rectangle (1,2) ;
    \fill [black!50] (1,3) rectangle (2,2) ;
    \fill [black!50] (3,2) rectangle (5,1) ;
    \fill [black!50] (6,0) rectangle (8,1) ;
    \fill [black!50] (11,11) rectangle (15,7) ;
    \fill [white] (12,10) rectangle (13,8) ;
    \fill [white] (14,10) rectangle (15,9) ;
    \fill [black!50] (8,8) rectangle (9,3) ;
    \fill [black!50] (9,7) rectangle (13,4) ;
    \fill [white] (9,6) rectangle (12,5) ;
    \fill [black!50] (13,6) rectangle (15,5) ;
    \fill [black!50] (10,3) rectangle (13,0) ;
    \fill [black!50] (13,3) rectangle (15,2) ;
    \fill [black!50] (14,2) rectangle (15,1) ;
    \fill [white] (11,2) rectangle (12,1) ;

 \foreach \i in {0,...,15}
   {\draw (\i, 0) -- +(0,15) ;
    \draw (0, \i) -- +(15,0) ;
   }
\end{tikzpicture}
\hspace{20pt}
\begin{tikzpicture} [scale = 0.15, baseline = 0.7cm]
 \fill [black!50] (0,15) rectangle (4,11) ;
 \fill [white] (1,14) rectangle (3,13) ;
  \fill [black!50] (5,15) rectangle (6,14) ;
    \fill [black!50] (9,15) rectangle (11,14) ;
    \fill [black!50] (9,14) rectangle (10,11) ;
    \fill [black!50] (10,12) rectangle (15,11) ;
    \fill [white] (12,12) rectangle (13,11) ;
    \fill [black!50] (14,15) rectangle (15,12) ;
    \fill [white] (14,14) rectangle (15,13) ;
    \fill [black!50] (3,11) rectangle (5,9) ;
    \fill [black!50] (7,10) rectangle (9,9) ;
    \fill [black!50] (0,10) rectangle (2,9) ;
    \fill [black!50] (1,9) rectangle (2,7) ;
    \fill [black!50] (3,6) rectangle (5,5) ;
    \fill [black!50] (4,9) rectangle (5,6) ;
    \fill [black!50] (0,4) rectangle (1,2) ;
    \fill [black!50] (1,3) rectangle (2,2) ;
    \fill [black!50] (3,2) rectangle (5,1) ;
    \fill [black!50] (6,0) rectangle (8,1) ;
    \fill [black!50] (11,11) rectangle (15,7) ;
    \fill [white] (12,10) rectangle (13,8) ;
    \fill [white] (14,10) rectangle (15,9) ;
    \fill [black!50] (8,8) rectangle (9,3) ;
    \fill [black!50] (9,7) rectangle (13,4) ;
    \fill [white] (9,6) rectangle (10,5) ;
    \fill [white] (11,6) rectangle (12,5) ;
    \fill [black!50] (13,6) rectangle (15,5) ;
    \fill [black!50] (10,3) rectangle (13,0) ;
    \fill [black!50] (13,3) rectangle (15,2) ;
    \fill [black!50] (14,2) rectangle (15,1) ;
    \fill [white] (11,2) rectangle (12,1) ;

 \foreach \i in {0,...,15}
   {\draw (\i, 0) -- +(0,15) ;
    \draw (0, \i) -- +(15,0) ;
   }
\end{tikzpicture}
\hspace{20pt}
\begin{tikzpicture} [scale = 0.15, baseline = 0.7cm]
 \fill [black!50] (0,15) rectangle (4,11) ;
 \fill [white] (1,14) rectangle (3,13) ;
  \fill [black!50] (5,15) rectangle (7,14) ;
    \fill [black!50] (6,14) rectangle (7,12) ;
    \fill [black!50] (9,15) rectangle (11,14) ;
    \fill [black!50] (9,14) rectangle (10,11) ;
    \fill [black!50] (10,12) rectangle (15,11) ;
    \fill [white] (12,12) rectangle (13,11) ;
    \fill [black!50] (14,15) rectangle (15,12) ;
    \fill [black!50] (3,11) rectangle (5,9) ;
    \fill [black!50] (5,10) rectangle (9,9) ;
    \fill [black!50] (0,10) rectangle (2,9) ;
    \fill [black!50] (1,9) rectangle (2,5) ;
    \fill [black!50] (2,6) rectangle (6,5) ;
    \fill [black!50] (4,9) rectangle (5,6) ;
    \fill [black!50] (0,4) rectangle (1,2) ;
    \fill [black!50] (1,3) rectangle (4,2) ;
    \fill [black!50] (3,2) rectangle (5,1) ;
    \fill [black!50] (6,4) rectangle (7,2) ;
    \fill [black!50] (6,0) rectangle (8,1) ;
    \fill [black!50] (11,11) rectangle (15,7) ;
    \fill [white] (12,10) rectangle (13,8) ;
    \fill [white] (14,10) rectangle (15,9) ;
    \fill [black!50] (8,8) rectangle (9,3) ;
    \fill [black!50] (9,7) rectangle (13,4) ;
    \fill [white] (9,6) rectangle (10,5) ;
    \fill [white] (11,6) rectangle (12,5) ;
    \fill [black!50] (13,6) rectangle (15,5) ;
    \fill [black!50] (10,3) rectangle (13,0) ;
    \fill [black!50] (13,3) rectangle (15,2) ;
    \fill [black!50] (14,2) rectangle (15,1) ;

 \foreach \i in {0,...,15}
   {\draw (\i, 0) -- +(0,15) ;
    \draw (0, \i) -- +(15,0) ;
   }
\end{tikzpicture}
\end{equation*}
\caption{Filtered digital image.}
\label{fig:filtered-image}
\end{figure}
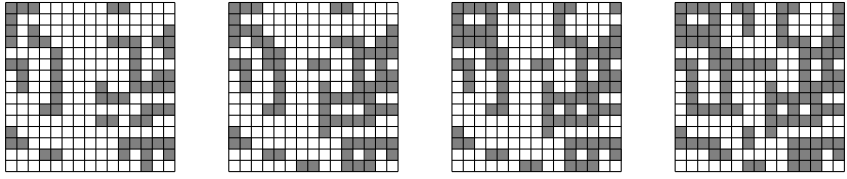

Let us consider for example the filtered image of Figure \ref{fig:filtered-image}. The bases of the corresponding simplicial complex are made of $203$ vertices, $408$
edges and $208$ triangles. Applying Theorem \ref{thm:image-dvf-persistenthomology} one can construct a reduction (compatible with the filtration) to a small chain complex which in this case has \emph{only} $16$ vertices, $20$ edges and $7$ triangles. The reduction is obtained in two steps (dimensions $1$ and $2$), by using two discrete vector fields (obtained by applying Theorem \ref{thm:matrix-dvf} to different submatrices) with respectively $187$ and $201$ vectors. The reduction provides an effective homology for the initial image allowing one to compute its persistent homology groups in a more efficient way.

The final homology groups of the image are $H_0=\Z^7$ and $H_1=\Z^4$. Making use of our programs for computing persistent homology groups (by means of DVFs), we can see the \emph{evolution} of the corresponding homology classes along the four filtration steps . For example, $H^{1,4}_0=\Z^4$, which means that in dimension $0$ there are $4$ classes which are born at the first step and are still alive at (the last) step $4$: \newpage
{\footnotesize \begin{verbatim}
> (prst-hmlg-group K 1 4 0)
Persistent Homology H^{1,4}_0
Component Z
Component Z
Component Z
Component Z
\end{verbatim}}

\normalsize Similarly, $H^{2,4}_1=\Z^2$ means that there are $2$ holes at stage $2$ which are still alive at step~$4$:
{\footnotesize \begin{verbatim}
> (prst-hmlg-group K 2 4 1)
Persistent Homology H^{2,4}_1
Component Z
Component Z
\end{verbatim}}

The \emph{toy} example of Figure \ref{fig:filtered-image} shows the improvement provided by the use of DVFs. This improvement is of course much more significant when working with big digital images. For instance, we have used our programs to compute the persistent homology of several \emph{fingerprints} extracted from the repository~\cite{FVC}. Given a fingerprint image, we can filter it by taking at the first step some initial horizontal lines, adding at each stage of the filtration some additional lines and ending with the whole image. This filtration produces some persistent homology groups. A similar process could be done in the vertical direction, taking successively the columns of the image, producing in that way different persistent homology groups. It could seem natural to think that given two (different) fingerprint images corresponding to the same person, the so obtained persistent homology groups should be similar.

\begin{figure}
\begin{tikzpicture}
\draw (-1,0) node{\includegraphics[scale=0.1]{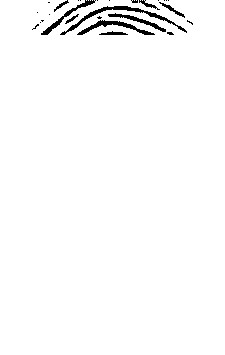}};
\end{tikzpicture}
\begin{tikzpicture}
\draw (-1,0) node{\includegraphics[scale=0.1]{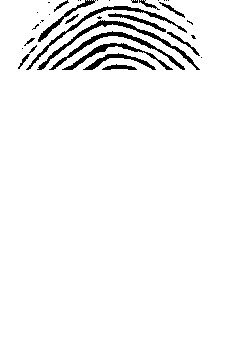}};
\end{tikzpicture}
\begin{tikzpicture}
\draw (-1,0) node{\includegraphics[scale=0.1]{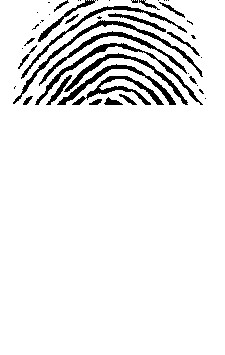}};
\end{tikzpicture}
\begin{tikzpicture}
\draw (-1,0) node{\includegraphics[scale=0.1]{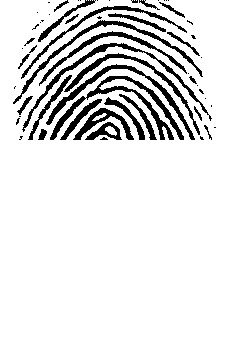}};
\end{tikzpicture}
\begin{tikzpicture}
\draw (-1,0) node{\includegraphics[scale=0.1]{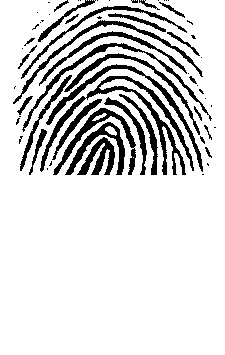}};
\end{tikzpicture}
\begin{tikzpicture}
\draw (-1,0) node{\includegraphics[scale=0.1]{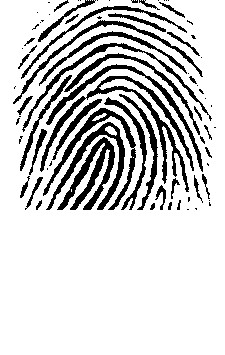}};
\end{tikzpicture}
\begin{tikzpicture}
\draw (-1,0) node{\includegraphics[scale=0.1]{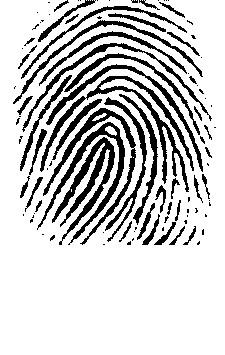}};
\end{tikzpicture}
\begin{tikzpicture}
\draw (-1,0) node{\includegraphics[scale=0.1]{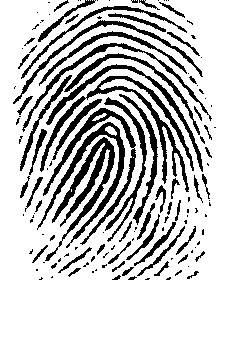}};
\end{tikzpicture}
\begin{tikzpicture}
\draw (-1,0) node{\includegraphics[scale=0.1]{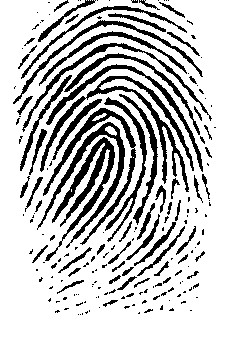}};
\end{tikzpicture}
\begin{tikzpicture}
\draw (-1,0) node{\includegraphics[scale=0.1]{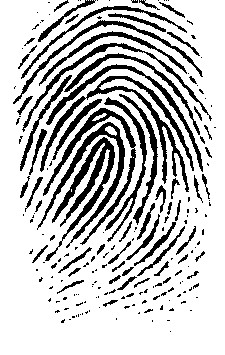}};
\end{tikzpicture}
\caption{Fingerprint filtration.}
\label{fig:fingerprint1}
\end{figure}

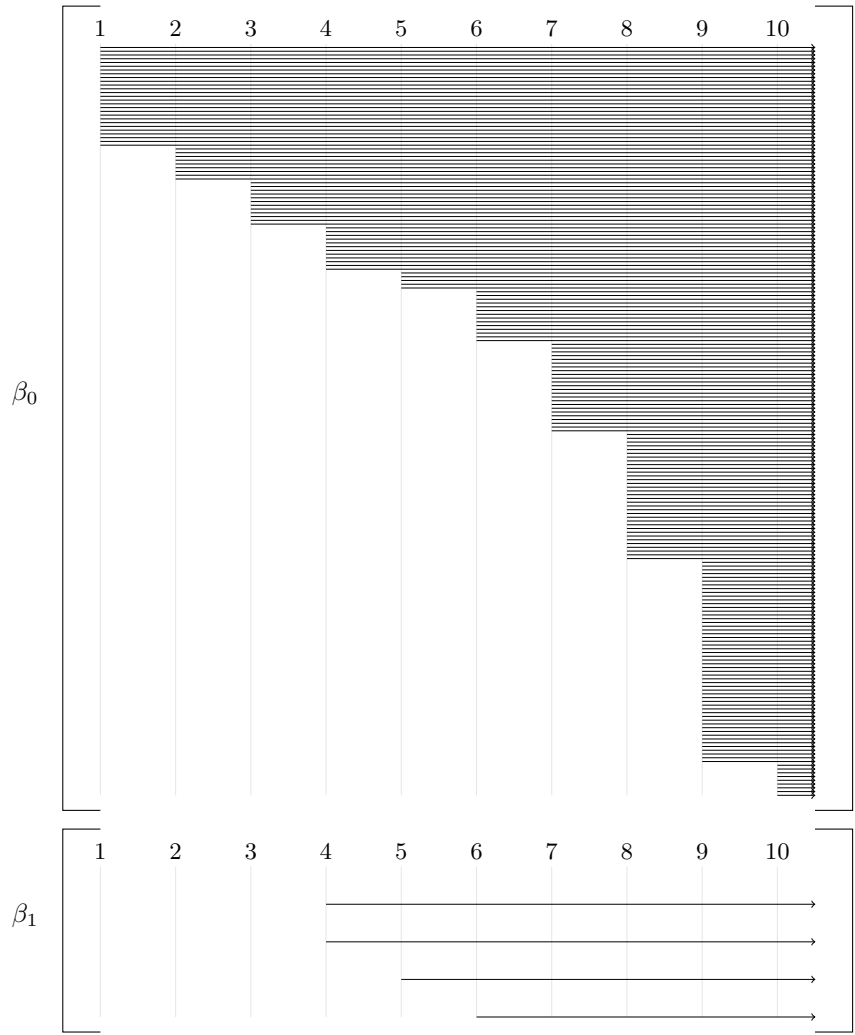
\begin{figure}
\begin{tikzpicture} [scale = 1, baseline = 0.5cm]
 \foreach \i in {1,2,...,10}
   {\draw[step=1,gray!20!white,very thin,fill=white] (\i, 0) -- (\i,10) ;
    \draw (\i,10.2) node {{\small \i}};
   }
  \foreach \i in {9.95,9.9,...,8.6}
   {\draw[->, ultra thin] (1,\i) -- (10.5,\i);
   }
   \foreach \i in {8.6,8.55,...,8.15}
   {\draw[->, ultra thin] (2,\i) -- (10.5,\i);
   }
   \foreach \i in {8.15,8.1,...,7.6}
   {\draw[->, ultra thin] (3,\i) -- (10.5,\i);
   }
   \foreach \i in {7.55,7.5,...,6.95}
   {\draw[->, ultra thin] (4,\i) -- (10.5,\i);
   }
   \foreach \i in {6.95,6.9,...,6.7}
   {\draw[->, ultra thin] (5,\i) -- (10.5,\i);
   }
   \foreach \i in {6.7,6.65,...,6}
   {\draw[->, ultra thin] (6,\i) -- (10.5,\i);
   }
   \foreach \i in {6,5.95,...,4.8}
   {\draw[->, ultra thin] (7,\i) -- (10.5,\i);
   }
   \foreach \i in {4.8,4.75,...,3.1}
   {\draw[->, ultra thin] (8,\i) -- (10.5,\i);
   }
   \foreach \i in {3.1,3.05,...,0.4}
   {\draw[->, ultra thin] (9,\i) -- (10.5,\i);
   }
   \foreach \i in {0.4,0.35,...,0}
   {\draw[->, ultra thin] (10,\i) -- (10.5,\i);
   }
   \draw (1,-0.2) -- (0.5,-0.2) -- (0.5,10.5) -- (1,10.5);
    \draw (10.5,-0.2) -- (11,-0.2) -- (11,10.5) -- (10.5,10.5);
    \draw (0,5.35) node {$\beta_0$};
 \end{tikzpicture}
 \vskip0.2cm
 \begin{tikzpicture} [scale = 1, baseline = 0.5cm]
 \foreach \i in {1,2,...,10}
   {\draw[step=1,gray!20!white,very thin,fill=white] (\i, 0) -- (\i,2) ;
    \draw (\i,2.2) node {{\small \i}};
   }
   \draw[->, ultra thin] (6,0) -- (10.5,0);
   \draw[->, ultra thin] (5,0.5) -- (10.5,0.5);
   \draw[->, ultra thin] (4,1) -- (10.5,1);
   \draw[->, ultra thin] (4,1.5) -- (10.5,1.5);
   \draw (1,-0.2) -- (0.5,-0.2) -- (0.5,2.5) -- (1,2.5);
    \draw (10.5,-0.2) -- (11,-0.2) -- (11,2.5) -- (10.5,2.5);
    \draw (0,1.35) node {$\beta_1$};
 \end{tikzpicture}
 \caption{Barcode of a fingerprint filtration.}
\label{fig:fingerprint-barcode}
\end{figure}

Let us consider for example the fingerprint of Figure \ref{fig:fingerprint1}, with ten steps horizontal filtration. The associated simplicial complex has $9082$ vertices, $20364$
edges and $11352$ triangles. The direct computation in Kenzo of its persistent homology groups is slow and in some cases fails because of memory problems. The discrete vector field reduction produces a \emph{small} chain complex with only $150$ vertices, $86$ edges and $6$ triangles, and all persistent homology groups can then be computed in less than a second. Figure \ref{fig:fingerprint-barcode} shows the barcode with the results computed by our programs. In this case, there are not deaths because all homology classes in both degrees $0$ and $1$ persist along the different stages of the filtration. Let us observe that for degree $1$, the two classes born at stage~$4$ correspond to real circles (holes) in the fingerprint; however, classes born at stages $5$ and $6$ are small holes produced in the image due to the resolution but not present in the fingerprint.

These good results have been also repeated for a number of fingerprints from the same source. Once the repository of fingerprints has been used to study the degree of reduction obtained by our programs, it could be natural to ask whether persistent homology could be usable to tackle the problem of \emph{fingerprint recognition}. After reflecting on this question, we are quite pessimistic with respect to this possibility. The main reason is that global properties of fingerprint pictures seem to depend more on the quality of the images (or, putting in other words, on the way a fingerprint has been acquired) than on the fact of corresponding to the same person. There is here a big problem with \emph{denoising} (erasing irrelevant features of the pictures). Of course, persistent homology has been applied to the problem of denoising by means of filtrations controlling the intensity or resolution (depending on the case of study) of images. Nevertheless, we believe that such an approach does not work on the case of fingerprints because of the same reason explained before.

Thus, we have tried another idea, trying to focus on \emph{local properties}. To this aim we have considered the layered filtration shown in Figure \ref{fig:fingerprint1} (this kind of filtration is not commonly used in the persistent homology setting), with the idea of tracking not only the homology groups, but also concrete cycles representing homology generators (recall that, even if the final computations are made on much smaller complexes, the cycles in the initial image are at our disposal thanks to its effective homology). The intuition guiding this approach is to relate essential cycles (at its birth index in the filtration) with some relevant points in fingerprints, called \emph{minutiae} in the fingerprint recognition literature (see, for instance, \cite{MMJP03}). Experimental evidence showing this idea could work is however weak, and we think that it would not be possible without heavy preprocessing steps or without applying selection algorithms in the whole datasets (based maybe on machine learning methods) prior to apply the persistent homology procedures. Taking into account ideas coming from differential geometry, one could also consider a foliation structure on a fingerprint. Foliation singularities could correspond to \emph{minutiae} in the fingerprint, which are relevant points to the recognition. In order to compute the different singularities, one could sweep the fingerprint in two different directions and study the corresponding barcode. 

\section{Conclusions and future work}

In this paper we have shown how the methods of effective homology can be applied to compute persistent homology of digital images. An algorithm to compute a discrete vector field (and the corresponding reduction to a smaller chain complex) for an image has been presented. Then, our algorithm has been unfolded to cover the case of a \emph{filtered} digital image, so allowing us to determine the persistent homology, together with the geometrical generators. These results generalise those presented in \cite{MN13}. Once implemented in the Kenzo system, our approach has shown a good reduction power both in artificial examples and in actual images extracted from a public fingerprints database. The usefulness of these homological techniques for the real problem of fingerprint recognition is still unclear and it is proposed as further work.

Other lines of research consist of looking for more efficient methods to compute DVFs for filtered images, and to take profit of the implicit parallelism of our algorithm. Finally, we could use our programs in other fields of application as the study of neuronal images \cite{HMPR11}.



\bibliographystyle{spmpsci}
\bibliography{biblio-ephdi}

A. Romero - J. Rubio. Department of Mathematics and Computer Science. University of La Rioja. Spain.
ana.romero@unirioja.es, julio.rubio@unirioja.es     

F. Sergeraert. Institut Fourier. University Joseph Fourier. France. \\
Francis.Sergeraert@ujf-grenoble.fr

\end{document}